\documentclass[10pt,twocolumn,letterpaper]{article}

\usepackage[pagenumbers]{cvpr} 
\usepackage{graphicx}
\usepackage{longtable}

\usepackage[rgb,table,xcdraw]{xcolor}
\usepackage{multirow}
\usepackage{arydshln}
\usepackage[most]{tcolorbox} 

\usepackage{adjustbox}
\definecolor{mygreen}{HTML}{3cb44b}
\definecolor{myred}{HTML}{800020}
\definecolor{myorange}{HTML}{FFBF00}
\definecolor{myblue}{HTML}{1f77b4}
\definecolor{mypurple}{HTML}{8000FF}

\newcommand{\PredSty}[1]{\textnormal{\ttfamily\color{mygreen!90!black}#1}\unskip}

\newcommand{\PredStyOrange}[1]{\textnormal{\ttfamily\color{myorange!90!black}#1}\unskip}

\usepackage{tabularx}
\usepackage[utf8]{inputenc}

\definecolor{cvprblue}{rgb}{0.21,0.49,0.74}
\usepackage[pagebackref,breaklinks,colorlinks,allcolors=cvprblue]{hyperref}

\definecolor{bluxtemp}{rgb}{0.2,0.4,0.8} 

\title{ReMatch: Boosting Representation through Matching for Multimodal Retrieval}

\author{
  Qianying Liu\textsuperscript{1,2\dag\textsection*} \
  Xiao Liang\textsuperscript{2*} \
  Zhiqiang Zhang\textsuperscript{2\ddag} \ Zhongfei Qing \and Fengfan Zhou\textsuperscript{3} \
  Yibo Chen\textsuperscript{2} \
  Xu Tang\textsuperscript{2} \
  Yao Hu\textsuperscript{2} \
  Paul Henderson\textsuperscript{1}  \and
  \textsuperscript{1}University of Glasgow, Glasgow, UK \ \textsuperscript{2}Xiaohongshu Inc., Beijing, China\\ \textsuperscript{3}Huazhong University of Science and Technology, Wuhan, China\and
\href{https://github.com/FireRedTeam/ReMatch}{\texttt{\textcolor{blue}{https://github.com/FireRedTeam/ReMatch}}}
}

\renewcommand{\paragraph}[1]{\vspace{2pt}\par\noindent\textbf{#1}~~}

\begin{document}
\maketitle
\begingroup
\renewcommand\thefootnote{\fnsymbol{footnote}}

\footnotetext[2]{Work was done during internship in Xiaohongshu Inc.}
\footnotetext[1]{Equal contribution}
\footnotetext[3]{Project lead}
\footnotetext[4]{Corresponding author (2665227L@student.gla.ac.uk)}

\endgroup
\begin{abstract}
We present \textbf{ReMatch}, a framework that leverages the generative strength of MLLMs for multimodal retrieval.
Previous approaches treated an MLLM as a simple encoder, ignoring its generative nature, and under-utilising its compositional reasoning and world knowledge.
We instead train the embedding MLLM end-to-end with a chat-style generative matching stage.
The matching stage uses the same MLLM to autoregressively decide relevance from multi-view inputs, including both raw data and its own projected embeddings for each query and document. It provides instance-wise discrimination supervision that complements a standard contrastive loss, offering stronger gradients on hard negatives and preserving the compositional strengths of the original MLLM.
To obtain semantically richer multimodal embeddings, we use multiple learnable tokens to augment each input, generating fine-grained contextual, mutually orthogonal embeddings with low inference cost.
Leveraging our established high-performance baseline,we assemble the ideas mentioned above into a powerful training recipe and achieve a new state-of-the-art on the Massive Multimodal Embedding Benchmark
(MMEB). 
Our experiments show particularly strong zero-shot generalization results on five datasets, highlighting the robustness and transferability of ReMatch.
\end{abstract}    
\section{Introduction}
\label{sec:intro}

\begin{figure}[t]
\begin{center}
\includegraphics[width=\linewidth]{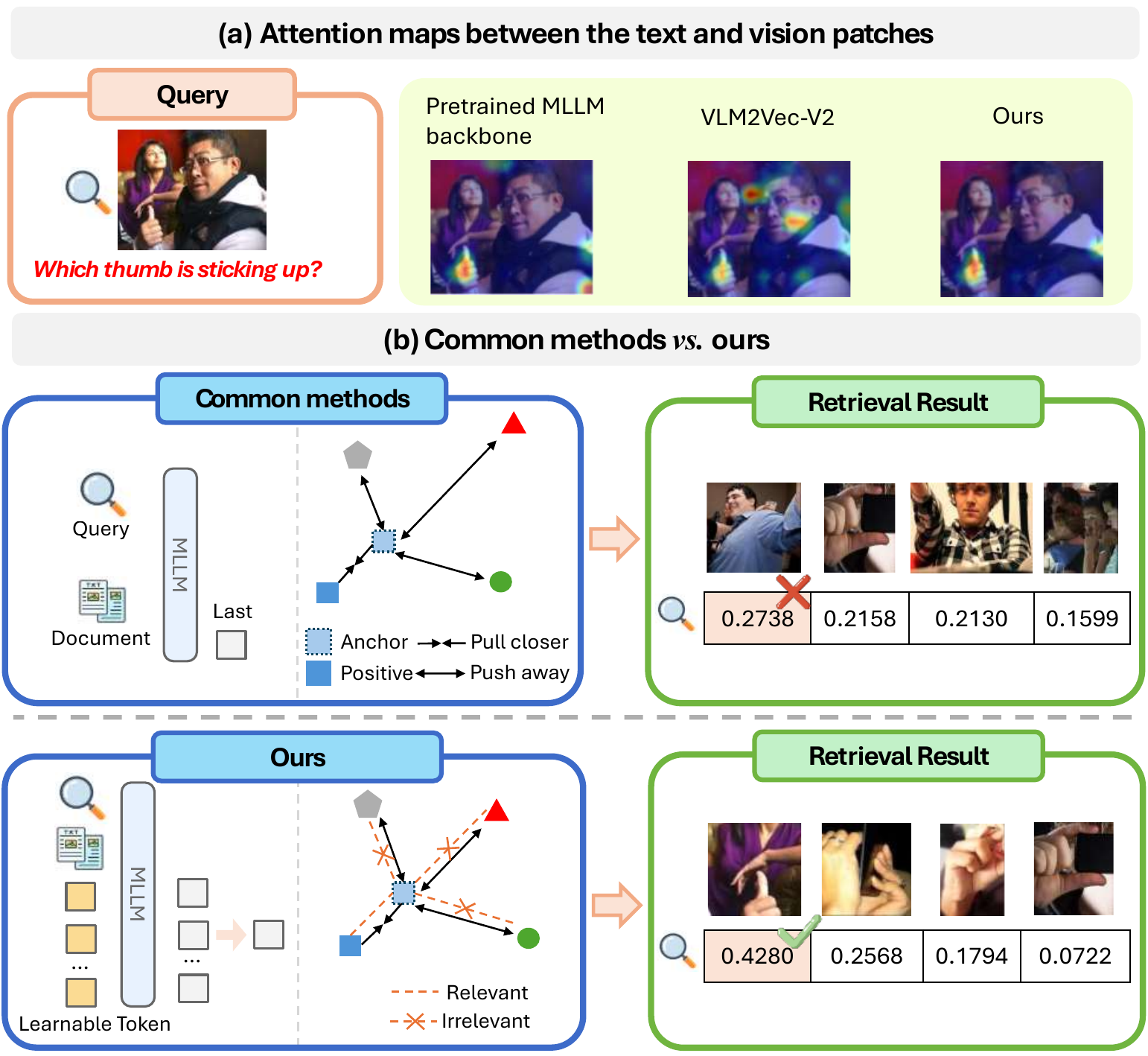}
    \end{center}
\caption{Illustration of our motivation. (a) Previous embedding methods like VLM2Vec-V2 disrupt the inherent fine-grained grounding of pretrained MLLMs. In contrast, our approach effectively preserves this critical alignment. (b) Our framework combines multiple learnable tokens with a generative matching objective to produce fine-grained and discriminative embeddings.}
\label{fig:motivation}
\end{figure}

Multimodal embedding models are fundamental to image search~\cite{gordo2017end}, visual question answering~\cite{hu2018learning, zhang2021multimodal}, and visual document retrieval~\cite{faysse2024colpali}. These models embed heterogeneous inputs—such as images and text—into a shared representation space, which allows us to solve tasks by finding matching embeddings between different modalities.  Approaches like CLIP~\cite{radford2021learning}, BLIP~\cite{li2022blip} and SigLIP~\cite{zhai2023sigmoid} show strong cross-modal retrieval performance, but still face limitations when inputs contain complex or highly varied instructions.

Recently, multi-modal large language models (MLLMs)  
have demonstrated strong potential for multimodal embedding, leveraging their native support for multimodal inputs~\cite{jiang2024e5, zhang2024gme, chen2025mme5, thirukovalluru2025breakingb3, zhou2025megapairs, cui2025think, li2025u}. Most existing approaches retain the causal‑attention paradigm of native MLLMs; using specially designed prompts~\cite{jiang2024e5} they then extract embeddings from the last layer hidden state, typically corresponding to either the last token or a special [EOS] token. The extracted embeddings are then used for contrastive learning rather than next-token prediction.

Despite substantial progress in embedding MLLMs, recent work~\cite{zhu2025freeret} shows that the final hidden state is optimized for generating vocabulary logits rather than preserving semantic structure, leading to suboptimal embeddings. Treating a MLLM as a single-vector embedding model without accounting for its autoregressive nature may reduce some of the pretrained backbone’s abilities, such as integrating information across multimodal regions of interest (see top of Figure~\ref{fig:motivation}). We hypothesize this for two main reasons.
Firstly, condensing the entire query and candidate into a single terminal vector (e.g.~the hidden state of an [EOS] token)
alters the model’s original generative usage, where such tokens primarily serve 
logits-predict roles, and can not capture semantically rich representations in their hidden states~\cite{zhu2025freeret}
. Moreover a single vector is limited in capacity and cannot retain fine-grained details, particularly for high-dimensional modalities such as images~\cite{yao2021filip}.
Secondly, relying solely on a global contrastive/discriminative loss may limit the preservation of its vision-language compositional understanding acquired during generative pretraining~\cite{ouali2025vladva}.

In this work, we propose \textit{ReMatch} (\textbf{R}epresentation \textbf{E}nhanced via \textbf{Match}ing), a novel approach to learning multimodal embeddings that leverages generative strengths of MLLMs.
During training, it incorporates a chat-style matching stage ingesting the embeddings themselves, and a multi-token embedding strategy (see bottom of Figure~\ref{fig:motivation}). 

Our first key contribution is to introduce a \textbf{chat-style matching mechanism that is applied when fine-tuning the MLLM for embedding}; this re-uses the same MLLM to autoregressively predict the relevance of multi-view input: both the raw data and embeddings of each query and document.
This approach regularizes the MLLM and provides an stronger learning signal, encouraging its embeddings to capture finer semantic cues and allowing them to be fed back into the model for deeper understanding. It complements the standard InfoNCE contrastive objective, enhancing instance-wise image-text discrimination while preserving the compositional strengths of the original backbone. Furthermore, it provides stronger gradients for hard negatives, thereby mitigating negative-sampling bias.

As a second contribution, we introduce a small number of special \textbf{learnable tokens appended to the input sequence of the query and candidate, whose last-layer hidden states serve as contextualized representations}. This differs from the standard approach of encoding each query or candidate into one vector associated with the last token.
To ensure these tokens capture complementary information and avoid collapsing into similar directions, we apply a soft orthogonality loss.

To minimize the computational cost of multi-relevance decisions and late multi-token interaction \cite{xiao2025metaembed}, we first design a unified attention masking scheme that fully exploits the causal properties of MLLMs, enabling efficient multi-view matching within a single computation, and fuse multiple token embeddings into a single representation preserving diversity of information.

We consolidate training techniques 
into a reproducible plain baseline that nearly matches the current SOTA, establishing a new high-quality baseline that raises the bar for multimodal retrieval. On top of this enhanced baseline, our novel methods deliver further substantial gains and establish new SOTA performance on the MMEB benchmark.

\section{Related Work}

\paragraph{Multimodal Representation Embedding.}
CLIP~\cite{radford2021learning} and subsequent models such as BLIP~\cite{li2022blip}, CoCa~\cite{yu2022coca}, and SigLIP~\cite{zhai2023sigmoid} have advanced vision–language pretraining by improving loss functions, data quality, and training strategies. However, most of these models still adopt a dual-encoder architecture, which constrains deeper cross-modal interactions.
Recent studies have explored the use of MLLMs for multimodal representation learning~\cite{kong2025modality, chen2025moca, gu2025breakingmodalitybarrieruniversal, ouali2025vladva, shin2025generative}. For instance, E5-V~\cite{jiang2024e5} employs specially designed prompts to generate embeddings, while VLM2Vec~\cite{jiang2024vlm2vec} introduces a contrastive learning framework in conjunction with the Massive Multimodal Embedding Benchmark (MMEB). GME~\cite{zhang2024gme} further synthesizes fused-modal training data to enable universal multimodal retrieval. Other approaches enhance retrieval performance by generating large-scale synthetic datasets~\cite{chen2025mme5, zhou2025megapairs} or leveraging hard negatives through advanced sampling strategies and gradient amplification~\cite{thirukovalluru2025breakingb3, qqmm, lan2025llave}. MetaEmbed~\cite{xiao2025metaembed} adopts learnable tokens to produce compact multi-vector representations for fine-grained cross-modal matching. However, it uses a ColBERT-style~\cite{khattab2020colbert} late-interaction paradigm, which introduces non-negligible computational overhead. In contrast, our approach retains the richness of multi-token embeddings while matching the computational efficiency of traditional single-vector methods.

\begin{figure*}[t]
\begin{center}
\includegraphics[width=1\linewidth]{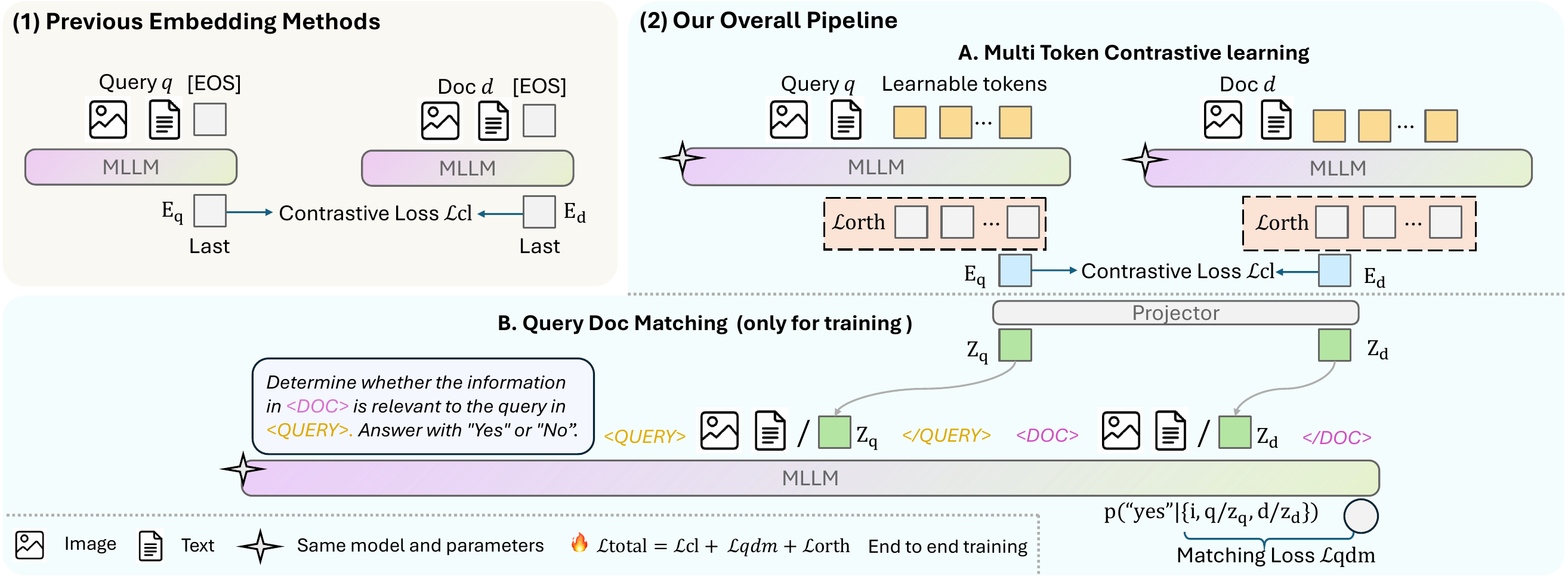}
    \end{center}
\caption{Previous multimodal retrieval frameworks \emph{v.s.} our \textit{ReMatch}. \textbf{Upper Left:} Single token retrieval method outputs an embedding for each pair of query and doc corresponding the [EOS] position, and uses contrastive objective to maximize the similarity for
corresponding pairs.
\textbf{Upper Right:} our framework first augments the input with Learnable Tokens and obtains multi-vector representations at these learnable-token positions. Then orthogonal regularization are employed on these representations and fuse into one embedding for every query or doc which are optimized by contrastive objective. The output embeddings are adapted by a MLP projector into MLLM input distribution, which used by our matching loss. 
\textbf{Lower:} we propose Query Doc Matching strategy to add point-wise discriminative 
 signals in framework from original input and feature perspectives.
}
\label{fig:framework}
\end{figure*}

\paragraph{Matching Module for Representation Learning.}
Early multimodal models such as ALBEF~\cite{li2021align} and BLIP~\cite{li2022blip} introduce an image–text matching module to capture fine-grained vision–language alignments. In the MLLM era, matching is typically treated as an independent stage following retrieval~\cite{liu2025lamra, gu2025unimev2, xu2025mm}, consistent with information retrieval paradigms. LamRA~\cite{liu2025lamra} introduces a two-stage framework where a separate reranker performs point-wise or list-wise reranking over results retrieved by LamRA-Ret to refine retrieval performance. UniME-V2~\cite{gu2025unimev2} leverages a MLLM as a judge to mine hard negatives for representation learning and to train a reranker with improved semantic discrimination. However, integrating such matching mechanisms into the MLLM paradigm to further enhance representation learning remains underexplored.

\section{Method}
\label{sec:Method}
We first (Sec.~\ref{Preliminaries}) revisit the definition of multimodal retrieval and outline our MLLM embedding baseline, discussing its key limitations that we aim to mitigate.
We then present the ReMatch recipe (Sec.~\ref{Our design}), and our main technical contributions (Sec.~\ref{sec:Learnable Multi-Token} and Sec.~\ref{sec:qdm}).

\subsection{Preliminaries}
\label{Preliminaries}

\paragraph{Problem Definition.} 
Multimodal retrieval involves retrieving relevant content across different modalities (e.g.~text, image, video). We focus on text–image retrieval setting where the query ${q}$ can be text ($q_t$), an image ($q_i$), or a combination of both $(q_t, q_i)$; each candidate document $d$ can similarly be any combination of text and image. 
Given a query $q$ and a set of $N$ candidate documents $\{d_1, d_2, \ldots, d_N\}$, a multimodal retrieval model defines the query embedding $E({q})$ and document embeddings $E({d})$, and then uses a similarity function $s\bigl(E(q),E(d)\bigr)$ to measure the relevance between $E({q})$ and $E({d})$. 
The top-1 retrieved result $d^\star$ is then determined by: 
\begin{equation}
d^\star = \mathop{\arg\max}\limits_{d \in \{d_1, \ldots, d_N\}} s\bigl(E(q), E(d)\bigr).
\end{equation}

\paragraph{Multimodal embedding large models.}
Recent retrieval methods project each modality to a shared embedding space via MLLM before measuring similarity.
Concretely, state-of-the-art (SOTA) multimodal embedding models \cite{chen2025mme5,jiang2024vlm2vec,zhang2024gme} are built on MLLMs with causal attention, typically appending an
[EOS] token at the end of the input sequence. Given a multimodal input (query or document)
$x$, where consists of both text tokens and image patches, and $x$ then is tokenized into a representation $m=[m_1,\dots,m_T]$. 
A decoder-only transformer produces a sequence of hidden states  $f_\theta(x)=f_\theta(m; [EOS]) \in \mathbb{R}^{(T+1) \times d}$, where $d$ is the hidden-state dimension and the $j$-th state only depends on
$[m_1, \dots, m_j]$. A natural choice for a single representation of $x$ is derived from the hidden state of the last layer corresponding to this [EOS] token \cite{zhang2024gme}: $\mathrm{E}_\theta(x):= (f_\theta(x))_{T+1}$.

\paragraph{Discussion.} 

The above framework raises two key concerns: (i) Although separator tokens (e.g., “.”, “\texttt{\textbackslash n}” or [EOS]) are often associated with representations that aggregate contextual information \cite{chen2024sepllm}, naively extracting their last-layer hidden states as representations 
may impair the inherent capability of MLLMs \cite{ouali2025vladva}. Moreover, a single token is may be insufficient to encode a large, compositional image–text input.
(ii) Using only a contrastive loss optimises embedding distances globally, but may fail to capture finer correspondences between modalities~\cite{zhong2022regionclip}.
Addressing these limitations is crucial for learning compact yet expressive representations that better capture fine-grained semantics through richer contextualization. We next describe how ReMatch is designed to tackle the above issues.

\subsection{Our design}
\label{Our design}

We propose ReMatch, a framework that introduces chat-style matching and learnable multi-token augmentation in an end-to-end fashion, designed to better activate and leverage the capabilities of the original MLLM for enhanced representation learning. (Fig.~\ref{fig:framework}).
To this end, each input is augmented with $K$ learnable tokens (Sec.~\ref{sec:Learnable Multi-Token}) whose directions are encouraged to be orthogonal.
The resulting multi-token embeddings are fused and optimized by a contrastive loss (Sec.~\ref{sec:contrastive}).
Subsequently, the \emph{raw} query–document pair \emph{together with} their embeddings into the same MLLM with a chat-style prompt, and train it to output “Yes/No”, providing instance-level discrimination supervision that complements the contrastive loss while preserving the compositional reasoning capabilities of the original MLLM (Sec.~\ref{sec:qdm}).  

\subsubsection{Multi Embedding Fusion}
\label{sec:Learnable Multi-Token}

\paragraph{Learnable multi-token augmentation.} 
To address the first issue in discussion, relying on the [EOS], we augment each input with $Learnable\ Tokens$ to better embed data, which are learned independently of the original MLLM vocabulary, ensuring their embeddings are decoupled from the model’s generative outputs: $L_q \in \mathbb{R}^{K \times d}$ for queries and $L_d \in \mathbb{R}^{K \times d}$.
The transformer then consumes
$z^{(0)} = [m; L_q; L_d] \in \mathbb{R}^{(T + K +K)\times d}$,
and produces last-layer hidden states
$H = f_\theta\bigl(z^{(0)}\bigr) \in \mathbb{R}^{(T + K)\times d}$.
We extract the final hidden states at these learnable-token positions to obtain a compact, contextualized multi-vector representation
$ME_q =[\,e_1, \dots, e_K\,] \in \mathbb{R}^{K \times d}$ and $\quad
ME_d \in \mathbb{R}^{K \times d}$, in two separate forward passes of the MLLM.

\paragraph{Orthogonal regularization.} 
Since the $K$ learnable tokens will later be fused into a single embedding and supervised by a contrastive loss, we would like these $K$ token-level embeddings to carry \emph{non-overlapping} information. 
We use a soft orthogonality constraint, replacing the hard constraint $e_i^\top e_j = 0$ with a differentiable penalty on pairwise inner products. 
Concretely, given the $K$ learnable-token embeddings of 
$E_q$ or $E_d =[\,e_1, \dots, e_K\,]$ ,
we first L2-normalize them along the feature dimension
$\tilde e_i = e_i / \lVert e_i \rVert_2$.
We then penalize the embedding pairs if $i\neq j$:

\begin{equation}
\label{eq:orth}
\mathcal{L}_{\text{orth}}
= \frac{2}{K(K-1)}
  \sum_{1 \le i < j \le K}
  \bigl( \tilde e_i^\top \tilde e_j \bigr)^2.
\end{equation}
In practice, we compute Eq. \ref{eq:orth} for every sample in the batch and take the mean, encouraging the fused representation to aggregate diverse cues.
After regularizing token diversity, we fuse the $K$ token embeddings by simple averaging:
$E_* = \frac{1}{K} \sum_{i=1}^K e_i \in \mathbb{R}^d$
where $*$ denotes $q$ or $d$, and use $(E_q, E_d)$ in the contrastive loss introduced in the following.

\subsubsection{Contrastive Fine-tuning} 
\label{sec:contrastive}
Each training instance is structured as a tuple
$\mathcal{B}_i = \bigl(q_i, d_i^{+}, d_{i}^{-}\bigr),$
where $q_i$ is the query, $d_i^{+}$ is its positive document, and
$d_{i}^{-}$ is explicit hard negative. Note that in contrastive learning, documents of other instances in the same batch, including both positive documents and negative documents, are also used as in-batch negatives, denoted by $\mathcal{B}$.
Denote by $E_{q_i}$ and $E_{d}$ the final embeddings of the query and documents, respectively.
The contrastive loss for $\mathcal{B}_i$ is then defined as
\begin{align}
\label{eq:contrastive}
\mathcal{L}_{\mathrm{cl}}(\theta; \mathcal{B}_i)
&= - \log
\frac{\Phi(E_{q_i}, E_{d_i^{+}})}
     {\Phi(E_{q_i}, E_{d_i^{+}}) + N},
\nonumber 
\\
N &= \Phi(E_{q_i}, E_{d_{i}^{-}}) + \sum_{d \in \mathcal{B}} \Phi(E_{q_i}, E_{d}).
\end{align}
where $\Phi(\cdot, \cdot)$ is a similarity function, e.g.
$\Phi(a, b) = \exp\bigl(\mathrm{cos}(a, b) / \tau \bigr),$
with $\mathrm{cos}(\cdot, \cdot)$ the cosine similarity and $\tau$ a temperature.

\subsubsection{Multi-View Query-Document Matching}
\label{sec:qdm}

Our MVQDM (Multi-View Query-Document Matching) strategy addresses the second challenge in the discussion of Sec.~\ref{Preliminaries}: improving fine-grained discriminative ability between samples. 
This strategy supplements InfoNCE \cite{oord2018representation} by injecting pointwise signals that drive absolute relevance judgments and deliver stronger gradients on hard negatives—mitigating negative‐sampling bias.

\paragraph{Generative chat-style matching.}
Prior work (e.g., the BLIP series \cite{li2022blip,li2023blip}) feeds raw query–document pairs into the MLLM and applies an external binary classifier on its output.  
In contrast, we train the MLLM end-to-end to autoregressively generate a “Yes”/“No” token indicating query-document relevance directly from the paired inputs \((q, d^+ / d^-)\).  
We optimize the supervised fine-tuning loss defined as:
\begin{equation}
\label{eq:match-loss-cvpr-simple}
\mathcal{L}_{\mathrm{qdm}} = - \log p(l \mid P(\tilde{q},\tilde{d})),
\end{equation}
where $p(\cdot\mid*)$ is the probability assigned by the MLLM, and the label \(l\) is ``Yes'' for relevant $d^+$ and ``No'' for irrelevant $d^-$. $\tilde{q}$ and $\tilde{d}$ denote the query and document, which can be either raw data or projected embeddings (details are provided in the following section).  
\paragraph{Multi-view input \& efficiency.}
\label{sec:Multi-view input}
We design a multi-view input scheme for the MLLM, leveraging both the raw data $(q, d)$ and their embeddings $(E_q, E_d)$, as previously defined. 
Each embedding $E_*$ is first transformed by a lightweight MLP:
$Z_* = \text{MLP}(E_*)$,
producing embeddings compatible with the input distribution of the same MLLM. 
The raw data and projected embeddings are then used by the MLLM to predict relevance. 
This encourages the model to capture both fine-grained signals from embeddings and global contextual information from raw data, while keeping the entire computation within a single model instance.

\begin{figure}[t]
\begin{center}
\includegraphics[width=0.6\linewidth]{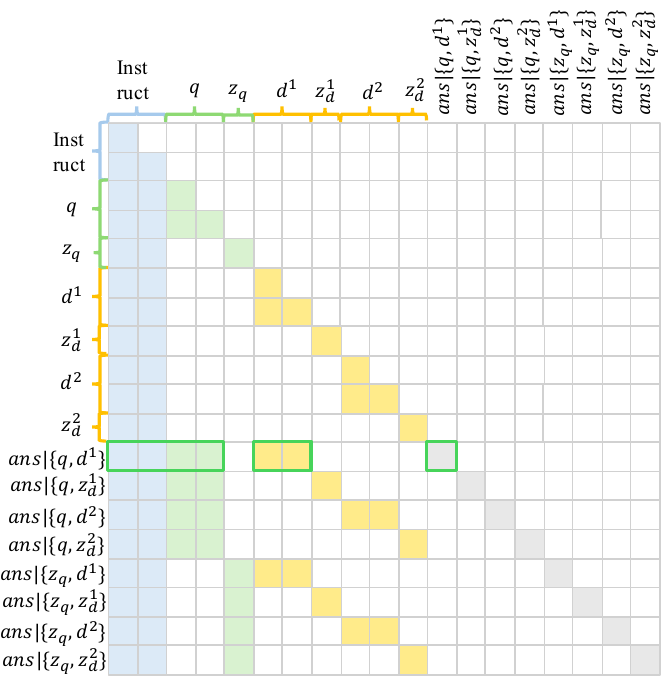}
    \end{center}
\caption{We introduce a unified attention mask that enables the model process all eight raw/embedding query–document combinations in one forward pass. Each answer token attends only to its paired query–document inputs (both raw and embedded) and the instruction prompt, preserving standard next-token prediction behavior. By randomizing which of $d^1,d^2$ holds the $d^+$, any positional leakage of relevance signals is prevented.}
\label{fig:attionmask}
\end{figure}

To efficiently handle all eight possible query-document views without multiple MLLM forward passes, we employ a unified attention mask (Figure~\ref{fig:attionmask}). 
Specifically, each query/document can be represented as raw data $q/d$ or as projected embeddings $Z_q/Z_d$. 
For each training instance, we sample one positive document $d^+$ and one negative document $d^-$, yielding $\{q,Z_q\}\times(\{d^+,Z_{d^+}\}+\{d^-,Z_{d^-}\})$ (2×(2+2)=8 combinations). 
To avoid positional bias, we allocate the two document slots \(d^1,d^2\) by randomly placing the positive document \(d^+\) in one slot and the negative document \(d^-\) in the other. We then train the model to output “Yes” when \(d^+\) is present and “No” otherwise.

Moreover, inspired by \cite{ouali2025vladva}, we adopt a conversational template for the matching process. This mirrors the MLLM’s generative pretraining regime, enabling us to fully leverage its pretrained generative capabilities.
Overall the MLLM receives:

\begin{minipage}{1.0\columnwidth}\vspace{2mm}\hspace{-7mm}    \centering
\begin{adjustbox}{width=1.0\columnwidth}
\begin{tcolorbox}
    \raggedright
    \small
    
\textbf{Prompt Template for Query–Document Matching:}\\
\textit{(Example shown for one query–document view.)}\\[1mm]

\texttt{USER: Compare the content inside } \PredStyOrange{\texttt{<QUERY>}} \texttt{ and } \textcolor{magenta}{\texttt{<DOC>}} \texttt{. Determine whether the information in } \textcolor{magenta}{\texttt{<DOC>}} \texttt{ is relevant to the query in } \PredStyOrange{\texttt{<QUERY>}} \texttt{. Answer only with "Yes" or "No".}\\

\PredStyOrange{\texttt{<QUERY>}}\PredSty{\texttt{\{query\}}/\texttt{\{query\_feat\}}}\PredStyOrange{\texttt{</QUERY>}} \\

\textcolor{magenta}{\texttt{<DOC>}}\PredSty{\texttt{\{doc\}}/\texttt{\{doc\_feat\}}}\textcolor{magenta}{\texttt{</DOC>}} \\

\texttt{ASSISTANT:} \PredStyOrange{\texttt{Yes/No}}

\vspace{-1mm}
\end{tcolorbox}
\end{adjustbox}
\vspace{1mm}
\end{minipage}

\paragraph{Lightweight alternative.}
To balance training complexity and model performance, we propose an efficient alternative that takes only the token‐level embeddings \((z_q, z_d)\) as input—omitting the raw multimodal data. Experimental results in Sec.~\ref{sec:abla:Matching} demonstrate that this lightweight scheme still yields significant retrieval improvements.

\subsubsection{Training Objectives}
Our framework is trained end-to-end, jointly optimizing both the contrastive and matching tasks.
The total loss combines the contrastive learning loss~\cite{oord2018representation}, orthogonality loss and query doc matching loss:
\begin{equation}
\label{eq:overall_train_obj}
\mathcal{L} = \mathcal{L}_{\text{cl}} + 
w_{\text{orth}} \mathcal{L}_{\text{orth}} + 
w_{\text{qdm}} \mathcal{L}_{\text{qdm}},
\end{equation}
where $w_\text{orth}$ and $w_\text{qdm}$ are set to constant values of $0.5$ and $0.1$ respectively.

\section{Experiments}
\label{sec:Experiment}

\begin{table*}[t]
  \caption{Hit@1 (\%) results on MMEB, which includes 36 tasks across four categories: Classification, Visual
Question Answering (VQA), Retrieval, and Visual Grounding. IND and OOD represent the in-domain average and
out-of-domain average metrics, respectively. \textbf{Bold} denotes the best scores in the subset and the second-best scores are
highlighted with \underline{underline}.} 
  \label{tab:main_result}
  \centering
  \scalebox{0.93}{
  \begin{tabular}{@{}lcccccccccc@{}}
    \toprule
    \multirow{2}{*}{\textbf{Model}} & \multirow{2}{*}{\textbf{Backbone}} & \multirow{2}{*}{\textbf{Size}} & \multicolumn{4}{c}{\textbf{Per Meta-Task Score}} & \multicolumn{3}{c}{\textbf{Average Score}} \\
    \cmidrule(lr){4-7} \cmidrule(l){8-10}
    & & & \textbf{Classification} & \textbf{VQA} & \textbf{Retrieval} & \textbf{Grounding} & \textbf{IND} & \textbf{OOD} & \textbf{Overall} \\
    \midrule
    \multicolumn{10}{c}{\cellcolor{gray!15}\textbf{\emph{Encoder-Based Baselines}}} \\
    CLIP \cite{radford2021learning} & ViT-L  & 428M & 55.2 & 19.7 & 53.2 & 62.2 & 47.6 & 42.8 & 45.4 \\
    UniIR \cite{wei2024uniir} & ViT-L14 & 428M & 44.3 & 16.2 & 61.8 & 65.3 & 47.1 & 41.7 & 44.7 \\
    MagicLens \cite{zhang2024magiclens} & ViT-L  & 613M & 38.8 & 8.3 & 35.4 & 26.0 & - & - & 27.8 \\
    \multicolumn{10}{c}{\cellcolor{gray!15}\textbf{\emph{MLLM-Based Baselines}}} \\
    VLM2Vec \cite{jiang2024vlm2vec} & Qwen2-VL & 2B &59.0 & 49.4 & 65.4 & 73.4 & 66.0 & 52.6 & 59.3 \\
    VLM2Vec-V2 \cite{meng2025vlm2vec} & Qwen2-VL  & 2B & 62.9 & 56.3 & 69.5 & 77.3 & – & – & 64.9 \\
    B3++ \cite{thirukovalluru2025breakingb3} & Qwen2-VL  & 2B & 67.0 & 61.2 & 70.9 & 79.9 & 72.1 & 63.1 & 68.1 \\
    \hdashline
    MoCa \cite{chen2025moca} & Qwen2.5-VL  & 3B & 59.8 & 62.9 & 70.6 & 88.6 & 72.3 & 61.5 & 67.5 \\
    \hdashline
    MM-EMBED \cite{lin2024mmembed} & LLaVa-Next  & 7B & 48.1 & 32.2 & 63.8 & 57.8 & – & – & 50.0 \\
    GME \cite{zhang2024gme} & Qwen2-VL  & 7B & 56.9 & 41.2 & 67.8 & 53.4 & – & – & 55.8 \\
    VLM2Vec \cite{jiang2024vlm2vec} & Qwen2-VL  & 7B & 61.2 & 49.9 & 67.4 & 86.1 & 67.5 & 57.1 & 62.9 \\
    MMRet \cite{zhou2025megapairs} & LLaVA-1.6 Mistral & 7B & 56.0 & 57.4 & 69.9 & 83.6 & 68.0 & 59.1 & 64.1 \\
    MoCa \cite{chen2025moca} & Qwen2.5-VL  & 7B & 65.8 & 64.7 & \textbf{75.0} & \underline{92.4} & 74.7 & 67.6 & 71.5 \\
    B3++ \cite{thirukovalluru2025breakingb3} & Qwen2-VL  & 7B & \textbf{70.0} & 66.5 & \underline{74.1} & 84.6 & 75.9 & 67.1 & 72.0 \\
    QQMM \cite{qqmm} & LLaVA-OneVision  & 7B & \underline{69.9} &70.0 &72.1 &86.0 &77.2& 66.6 &72.5 \\
    \hdashline
    mmE5 \cite{chen2025mme5} & Llama-3.2-Vision & 11B & 67.6 & 62.7 & 71.0 & 89.7 & 72.4 & 66.6 & 69.8 \\
    \midrule
    \multicolumn{10}{c}{\cellcolor{gray!15}\textbf{\emph{ReMatch - Qwen2-VL Initialized}}} \\
    ReMatch & Qwen2-VL  & 2B &65.8&65.9&70.1&83.3 &72.8 &64.7 &69.2\\
    ReMatch & Qwen2-VL  & 7B & 68.2& \underline{71.8} &73.9& 88.4&\underline{77.4} &\bf{68.3} &\underline{73.3} \\
    \hline
    \multicolumn{10}{c}{\cellcolor{gray!15}\textbf{\emph{ReMatch - Qwen2.5-VL Initialized}}} \\
    ReMatch & Qwen2.5-VL  & 3B & 62.4&69.6&70.0&92.0 &75.1 &64.0 &70.2\\
    ReMatch & Qwen2.5-VL  & 7B &65.8&\textbf{73.6}&\underline{74.1}&\textbf{92.5}&\textbf{78.1} &\underline{68.2} &\textbf{73.7} \\

    \bottomrule
  \end{tabular}
}
\end{table*}

\paragraph{Evaluation.}
To comprehensively evaluate the quality of method, we use the MMEB benchmark~\cite{jiang2024vlm2vec}\cite{meng2025vlm2vecv2advancingmultimodalembedding}. MMEB provides a comprehensive suite of 36 tasks to assess general multimodal capabilities, spanning four key areas: classification (CLS, 10 tasks), visual question answering (VQA, 10 tasks), retrieval (RET, 12 tasks), and visual grounding (VG, 4 tasks). Following the standard protocol, we report $Hit@1$ as the evaluation metric across all tasks.
In addition, to rigorously evaluate the efficacy of ReMatch's unimodal embeddings, we benchmark its performance on five cross-modal retrieval tasks. These tasks encompass short-caption retrieval on Flickr30K \cite{young-etal-2014-image} and COCO 2014 \cite{lin2015microsoftcococommonobjects}, long-caption retrieval on ShareGPT4V \cite{chen2023sharegpt4vimprovinglargemultimodal} and Urban1K \cite{zhang2024longclipunlockinglongtextcapability}, and compositional retrieval on SugarCrepe \cite{hsieh2023sugarcrepefixinghackablebenchmarks}.

 \paragraph{Implementation details.}
 We train ReMatch on 8 NVIDIA H800 GPUs for 2500 steps with a global batch size of 1024 to ensure fair comparison with previous methods \cite{meng2025vlm2vec, thirukovalluru2025breakingb3}. All models are fine-tuned using LoRA \cite{hu2022lora} with $r = 32$ and $\alpha = 64$. We adopt a learning rate of $10^{-4}$ with a cosine decay schedule. 
The contrastive temperature is fixed at $0.02$.
 Following \cite{thirukovalluru2025breakingb3,xue2025improve}, we also introduce positive target prompts to disentangle different tasks, stabilizing the training. As for training dataset, we only incorporate \textbf{MMEB-train} and  one explicit hard negative from~\cite{chen2025mme5} for training all variants of ReMatch.
 
\paragraph{Models.} To evaluate the effectiveness of ReMatch, we conduct experiments on
 various MLLMs of different sizes, including Qwen2-VL~\cite{wang2024qwen2vlenhancingvisionlanguagemodels}, 
 and Qwen2.5-VL~\cite{bai2025qwen25vltechnicalreport}. 
 They represent unified multimodal architectures that process text and vision inputs. Our approach is benchmarked against a diverse set of baselines. These include dual-encoder frameworks like CLIP \cite{radford2021learning}, UniLR \cite{wei2024uniir}, and MagicLens \cite{zhang2024magiclens}. We also conduct direct comparisons with VLM2Vec-V1 \cite{jiang2024vlm2vec} and VLM2Vec-V2\cite{meng2025vlm2vecv2advancingmultimodalembedding}, both of which leverage Qwen2-VL as their foundational MLLM. Furthermore, we assess performance against SOTA MLLM-based embedding techniques, such as MoCa \cite{chen2025moca}, MMRet \cite{zhou2025megapairs}, MM-EMBED \cite{lin2024mmembed}, GME \cite{zhang2024gme}, mmE5 \cite{chen2025mme5}, QQMM \cite{qqmm}, and B3++ \cite{thirukovalluru2025breakingb3}.

\subsection{Comparison with Existing Methods}

We report the overall multimodal embedding performance of different ReMatch variants and baseline methods on MMEB in Table~\ref{tab:main_result}. All ReMatch results are reported with 8 matching views and 16 learnable tokens; we measure the influence of these parameters in Sec.~\ref{sec:abla:Matching}.

The encoder-based baselines are evaluated in a zero-shot setting, while the remaining models are trained on MMEB. 
ReMatch achieves the best results at comparable model sizes, 
outperforming the next best approach by 1.1\% (B3++-2B), 2.7\% (MoCa-3B), 1.2\% (QQMM-7B). Compared to our baseline VLM2Vec, ReMatch achieves substantial improvements of 9.9\% and 10.4\%, respectively, on the 2B and 7B Qwen2-VL backbone.
Notably, on the VQA task, our ReMatch models achieve substantial gains (+4.7\% for the 2B groups and +3.6\% for the 7B groups), as it more effectively uses the MLLM’s latent world knowledge.
The choice of backbone has a pronounced effect on ReMatch performance across tasks. ReMatch with Qwen2-VL shows strong classification abilities (+3\%), but its VQA and grounding score drops sharply, more than 3 and 8 points lower than the Qwen2.5-VL-initialized model. We notice that if the underlying base model itself struggles on some domains when used as a generative model, such a weakness directly propagates into ReMatch as an embedding model, similar to the findings in \cite{huang2025mimicking}.
Moreover, ReMatch boosts retrieval quality without slowing down inference. Since relevance prediction occurs only during training, at test time ReMatch merely processes 16 additional tokens, incurring computational overhead comparable to VLM2Vec.

\subsection{Zero‐Shot Cross‐Modal Retrieval on Short \& Long Captions}

\begin{table*}[t]
\centering
\caption{Zero-shot text-image retrieval results (Recall@1) on short caption (Flickr30K, MS-COCO), long caption (ShareGPT4V, Urban1K), and compositional (SugarCrepe) datasets. }
\label{tab:fewshot}
\scalebox{0.83}{
\begin{tabular}{l c cc cc cc cc ccc}
\toprule
\multirow{3}{*}{\textbf{Models}} & \multirow{3}{*}{\textbf{Size}}
    & \multicolumn{4}{c}{\textbf{Short Caption}}
    & \multicolumn{4}{c}{\textbf{Long Caption}}
    & \multicolumn{3}{c}{\textbf{Compositional}} \\
    \cmidrule(r){3-6}\cmidrule(r){7-10}\cmidrule(r){11-13}
& & \multicolumn{2}{c}{\textbf{Flickr30K}} & \multicolumn{2}{c}{\textbf{COCO}}
  & \multicolumn{2}{c}{\textbf{ShareGPT4V}} & \multicolumn{2}{c}{\textbf{Urban1K}}
  & \multicolumn{3}{c}{\textbf{SugarCrepe}} \\
  \cmidrule(r){3-4} \cmidrule(r){5-6}\cmidrule(r){7-8}\cmidrule(r){9-10} \cmidrule(r){11-13}
  
& & $q^{t}\!\to\!c^{i}$ & $q^{i}\!\to\!c^{t}$ & $q^{t}\!\to\!c^{i}$ & $q^{i}\!\to\!c^{t}$
  & $q^{t}\!\to\!c^{i}$ & $q^{i}\!\to\!c^{t}$ & $q^{t}\!\to\!c^{i}$ & $q^{i}\!\to\!c^{t}$
  & Replace & Swap & Add \\
\midrule
OpenCLIP (ViT-L) \cite{Cherti_2023} & 0.4B
  & 67.3 & 87.2 & 37.0 & 58.1 & 81.8 & 84.0 & 47.0 & 47.0 & 79.5 & 62.7 & 74.9 \\
CLIP (ViT-BigG/14) \cite{radford2021learning} & 2.5B
  & 79.5 & 92.9 & 51.3 & 67.3 & 90.1 & 93.6 & 77.8 & 80.7 & 86.5 & 68.9 & 88.4 \\
EVA-CLIP \cite{sun2023eva} & 8B
  & 80.3 & 94.5 & 52.0 & 70.1 & 93.1 & 91.2 & 80.4 & 77.8 & 85.9 & 70.3 & 86.7 \\
\hdashline
VLM2Vec \cite{jiang2024vlm2vec} (Qwen2-VL) & 2B
  & 69.3 & 89.6 & 40.0 & 62.5 & 78.1 & 88.2 & 78.7 & 83.9 & 67.2 & 46.5 & 66.4 \\
UniME \cite{gu2025breakingmodalitybarrieruniversal} (Qwen2-VL) & 2B
  & 74.9 & 90.6 & 44.0 & 63.5 & 83.6 & 88.6 & 83.3 & 83.2 & 65.6 & 45.2 & 65.7 \\
UniME-V2 \cite{gu2025unimev2} (Qwen2-VL) & 2B
  & 79.8 & 89.9 & 53.7 & 65.1
  & 91.6 & \underline{94.2} & 95.6 & 92.2
  & 70.9 & 51.2 & 70.2 \\
\hdashline
E5-V \cite{jiang2024e5} (Phi3.5-V) & 4.2B
  & 72.2 & 79.6 & 44.7 & 53.4 & 86.0 & 88.5 & 83.8 & 83.6 & 88.2 & 66.6 & 75.3 \\
\hdashline
E5-V \cite{jiang2024e5} (LLaVA-1.6) & 7B
  & 77.3 & 85.7 & 49.1 & 57.6 & 85.1 & 82.1 & 88.9 & 83.2 & 86.3 & 68.7 & 66.9 \\
UniME \cite{gu2025breakingmodalitybarrieruniversal} (Qwen2-VL) & 7B
  & 80.8 & 92.7 & 50.9 & 69.8 & 86.5 & 93.8 & 95.3 & 94.0 & 68.8 & 53.0 & 69.8 \\
VLM2Vec \cite{jiang2024vlm2vec} (Qwen2-VL) & 7B
  & 80.0 & 94.2 & 49.2 & 68.5 & 78.5 & 90.4 & 94.0 & 94.2 & 70.0 & 51.7 & 72.2 \\
UniME \cite{gu2025breakingmodalitybarrieruniversal} (LLaVA-OV) & 7B
  & 83.3 & 94.4 & 54.8 & 74.0 & 93.9 & 89.3 & 94.3 & 95.5 & 80.5 & 65.5 & 82.2 \\
UniME-V2 \cite{gu2025unimev2} (Qwen2-VL) & 7B
  & 84.6 & 93.5 & 57.3 & 70.3
  & \underline{94.3} & \textbf{95.2} & \underline{97.2} & 96.3
  & 77.8 & 62.2 & 79.0 \\
UniME-V2 \cite{gu2025unimev2} (LLaVA-OV) & 7B
  & \underline{85.5} & 93.7 & \underline{60.9} & 74.1
  & \textbf{95.1} & 94.1 & 96.3 & 96.7
  & 88.6 & 73.7 & \underline{90.5} \\
\midrule
    \multicolumn{13}{c}{\cellcolor{gray!15}\textbf{\emph{ReMatch - Qwen2-VL Initialized}}} \\
ReMatch (Qwen2-VL) & 2B
  & 82.4 & \underline{94.6} & 56.8 & \underline{76.5} & 93.3 & 92.4 & 96.0 & \underline{96.9} & \underline{90.3} & \underline{73.8} & 90.3 \\
  ReMatch (Qwen2-VL) & 7B
  &  \textbf{85.6} & \textbf{95.4} & \textbf{62.8} & \textbf{79.3} & 91.0 & 90.8 & \textbf{97.5} & \textbf{98.3} & \textbf{91.5} & \textbf{77.7} & \textbf{93.9} \\
\bottomrule
\end{tabular}
} 

\end{table*}

We also evaluate ReMatch in a zero‐shot setting across three kinds of retrieval benchmarks (short‐caption; long-caption; compositional), underscoring the effectiveness of multi-token embedding paired with chat-style matching. As shown in Table \ref{tab:fewshot}, ReMatch-Qwen2-VL 2B delivers substantial gains over the strongest 2B‐scale baseline (UniME-V2): +2.6\% on text-to-image and +4.7\% on image-to-text in Flickr30K, and +3.1\%/ +11.4\% on COCO. On long‐caption tasks, ReMatch-2B exceeds UniME-V2 on Urban1K image-to-text by 4.7\%, demonstrating its robustness to richer descriptions. Remarkably, on the SugarCrepe object-swap task, ReMatch-2B  achieves +27.3\% gain over VLM2Vec 2B. Scaling to 7B further boosts performance: ReMatch sets new bests on four of the five benchmarks (Flickr30K, COCO, Urban1K, SugarCrepe), and, notably, outperforms UniME-V2 with Qwen2-VL-7B (79.0\%) by 14.9\% on the attribution-adding subset of SugarCrepe.

\subsection{Ablation Studies}

We conduct various ablation experiments on different tasks 
within MMEB, using our ReMatch-2B (Qwen2-VL).  We also show t-SNE visualization in the Appendix to compare the embedding distributions of our method and baselines.

\begin{table}[t]
\centering
\setlength{\tabcolsep}{0.8pt} 
\caption{Ablation study for the primary components of ReMatch 2B (Qwen2-VL). Training tuning: applies LoRA(rank=32, alpha=64), LR=1e-4 with cosine LR schedule, temperature=0.02 in CL loss on the baseline. MVQDM: multi-view query–document matching. MVQDM\textsuperscript{++}: further enhanced by incorporating a chat-style template into the embeding extraction phase to align with matching. MEF: multi embedding fusion. }
\label{tab:ablation}
\begin{tabular}{@{}l l c c c c c@{}}
\toprule
\small
\textbf{Exp} & \textbf{Configuration} & \textbf{Overall}\quad  & \textbf{CLS} & \textbf{QA} & \textbf{RET} & \textbf{VG} \\
\midrule
0  & Baseline (VLM2Vec)              &   59.7 \quad\quad\quad  &  58.7 &  49.3 &  65.0 &  72.9 \\
\hline
 \multicolumn{7}{c}{\cellcolor{gray!15}\emph{Our tuning variant}}           \\
1  & Training Tuning                  &   65.5 \textcolor{green}{$\uparrow 5.8$}  &  64.1 &  58.4 &  68.1 &  79.0 \\
2  & + Target Instruction             &   67.3 \textcolor{green}{$\uparrow 1.8$}  &  65.0 &  63.2 &  68.3 &  80.6 \\
3  & + Hard Negative Mining                    &   67.9 \textcolor{green}{$\uparrow 0.6$}  &  65.0 &  64.8 &  69.0 &  80.2 \\
\hline
\multicolumn{7}{c}{\cellcolor{gray!15}\emph{Our primary contributions}}        \\
4  & Exp3 + MVQDM  \                        &   68.4 \textcolor{green}{$\uparrow 0.5$}   &  65.5 &  64.8 &  69.6 &  80.8 \\
5  & Exp3 + MVQDM\textsuperscript{++}              &   68.8 \textcolor{green}{$\uparrow 0.9$}   & 65.7 & 65.2 & \textbf{70.2} & 81.4 \\
6  & Exp3 + MEF&   68.7 \textcolor{green}{$\uparrow 0.8$} & \bf{65.8 } &  65.2 &  70.1 &  80.6 \\

\midrule
7  & ReMatch-2B                   & \bf{69.2} \textcolor{green}{$\uparrow 9.5$}  &  \bf{65.8 }& \bf{65.9} & 70.1 & \bf{83.3} \\
\bottomrule
\end{tabular}
\end{table}

\paragraph{Benefit of components.}
We first measure the benefit of each of the primary components of ReMatch; 
results are shown in Table~\ref{tab:ablation}. 
Our baseline (VLM2Vec) achieves 59.7\% (top row). We first refine the training setup by tuning the parameters, adding task-specific target instructions and one explicit hard negative, yielding a strong tuning baseline that outperforms the original by 8.2\%. 
Next, we evaluate our main contributions. First, our Multi-View Query–Document Matching (MVQDM) strategy alone lifts performance by 0.5\%. To better align with the chat-style matching stage, we extend the embedding extraction with a chat-based prompt, denoted as MVQDM\textsuperscript{++}, which improves over our tuning baseline by 1.1\%.
Furthermore, adding our Multi Embedding Fusion (MEF) alone brings a 0.8\% improvement over tuning baseline.
Lastly, combing all novel components yields the full model's SOTA performance (69.2\%), demonstrating that both chat style MVQDM and MEF are crucial for enhancing performance.

\begin{table}[t]
\centering
\begin{minipage}[t]{0.2\textwidth}
  \centering
  \small
  \caption{Effect of weight factor $w_{\text{qdm}}$ for MVQDM loss.}
  \label{tab:match_weight}
  \begin{tabular}{@{}c c@{}}
  
    \toprule
     $w_{\text{qdm}}$ & \textbf{Overall} \\
    \midrule
    0.1  & \bf{68.4} \\
    0.2  & 68.2 \\
    0.4  & 68.0 \\
    0.8  & 68.1 \\
    \bottomrule
  \end{tabular}
\end{minipage}%
\hfill
\begin{minipage}[t]{0.25\textwidth}
  \centering
  \small
  \caption{Comparison of different matching views for MVQDM loss.}
  \label{tab:match_view}
  \begin{tabular}{@{}l c@{}}
    \toprule
    \textbf{Matching View}            & \textbf{Overall} \\
    \midrule
    Raw/Feat $\leftrightarrow$ Raw/Feat   & \bf{68.4} \\
    Feat $\leftrightarrow$ Feat               & 68.1 \\
    Raw $\leftrightarrow$ Raw                 & 68.2 \\
    Raw $\leftrightarrow$ Feat                & 68.2 \\
    Feat $\leftrightarrow$ Raw                & 68.1 \\
    \bottomrule
  \end{tabular}
\end{minipage}
\end{table}

\paragraph{Effectiveness of matching.}
\label{sec:abla:Matching}
To validate multi‐view strategy in MVQDM, we compare five matching configurations in Table~\ref{tab:match_view}. “Raw$\leftrightarrow$Raw” uses the original query–document pair $(q,d)$ as input, while “Feat$\leftrightarrow$Feat” uses their embedding pair $(z^q,z^d)$. Although any single view yields gains, the “Raw/Feat$\leftrightarrow$Raw/Feat” combination outperforms all others by a substantial margin (+0.5\% over Exp3 in Table~\ref{tab:ablation}), demonstrating that multi‐view matching delivers both finer‐grained discrimination and a powerful regularization effect. Notably, the Feat$\leftrightarrow$Feat view alone achieves performance nearly with Raw$\leftrightarrow$Raw (only 0.1\% lower), providing a lightweight matching solution and confirming that ReMatch’s projected embeddings are highly informative and could be readily understood by the same MLLM.

Table~\ref{tab:match_weight} reports the impact of the matching‐loss weight \(w_{\mathrm{qdm}}\) in Eq.~\eqref{eq:overall_train_obj}. We sweep \(w_{\mathrm{qdm}}\) from 0.1 to 0.8 and observe consistent gains of \(+0.2\)–\(+0.5\)\% over the contrastive‐only baseline (Exp3 in Table~\ref{tab:ablation}), confirming the robustness of our MVQDM strategy as an auxiliary loss. Notably, when combined with the chat template (Exp5 in Table~\ref{tab:ablation}), the matching yields up to a \(+0.9\%\) gain, illustrating its strong synergy with the generative matching mechanism.  

\begin{figure}[t]
\begin{center}
\includegraphics[width=0.92\linewidth]{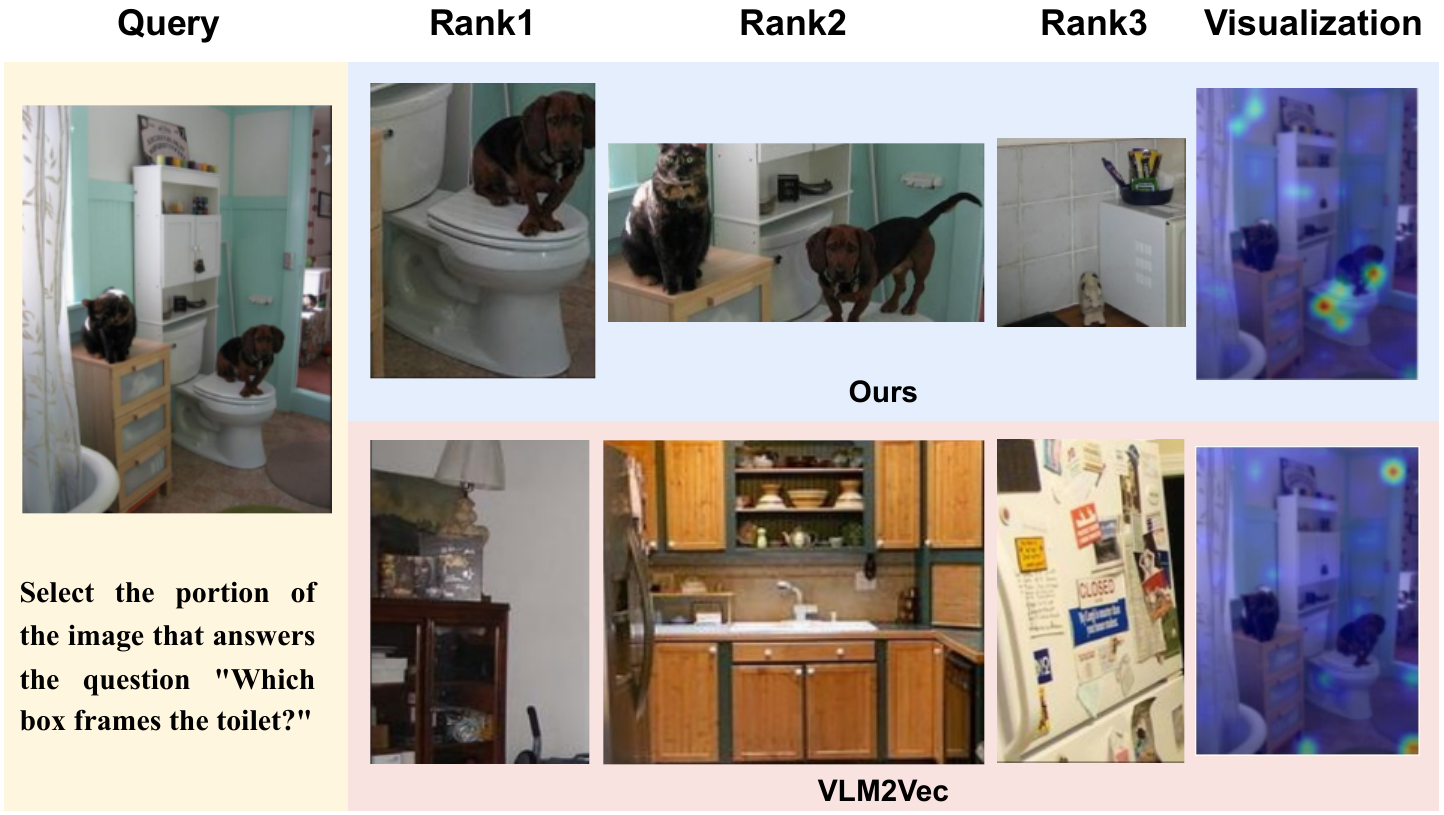}
    \end{center}
\caption{Qualitative Comparison of Visual Grounding Results on the Visual7W-Pointing (OOD) Dataset.
}
\label{fig:match_figure}
\end{figure}

Qualitative results are shown in Figure \ref{fig:match_figure}. Given an input image and a question, we compare the top-3 retrieved regions and attention maps of the final layer produced by our method with MVQDM\textsuperscript{++} (Exp5 in Table~\ref{tab:ablation}) and VLM2Vec-V2 (bottom row). Our method accurately retrieves the target patches, and the corresponding attention maps show focused activation on that interested object. This highlights a strong alignment between visual cues and the question semantics, demonstrating the fine‐grained multimodal discriminative ability of our MVQDM\textsuperscript{++}.

\paragraph{Effectiveness of multi-token augmentation.}
\label{sec:abla:Multi-Token}
Table~\ref{tab:mt-fuse} evaluates different token counts for our Multi Embedding Fusion (MEF). The first row corresponds to the single‐vector baseline using only the [EOS] hidden state (Exp3 in Table~\ref{tab:ablation}). Simply averaging $K$ learnable tokens yields consistent but fluctuating gains across $K \in \{4, 8, 16, 32, 64\}$, peaking at +0.5\%. When combined with our Orthogonal Multi-token Regularization (OR), the improvements become both larger and more stable, rising monotonically from +0.4\% at $K=4$ to +0.9\% at $K=64$, thereby demonstrating MEF’s effectiveness and scalability.

\begin{figure}[t]
\begin{center}
\includegraphics[width=0.9\linewidth]{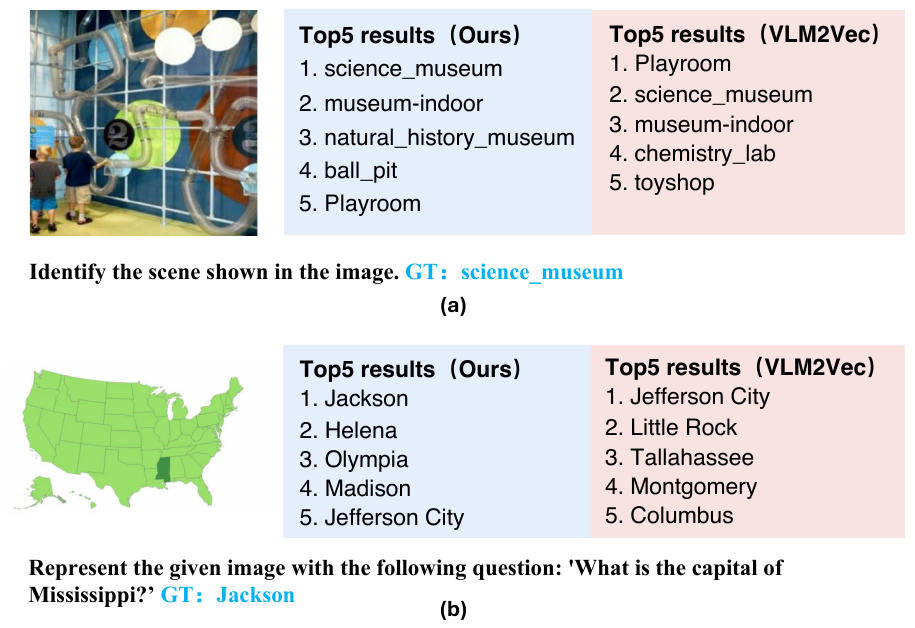}
    \end{center}
\caption{
VQA retrieval Results on Place365 (OOD) and ScienceQA (OOD).
Comparison between ours Exp6 and the baseline. 
}
\label{fig:multi-token}
\end{figure}

\begin{table}[t]
\centering
\small
\caption{Effect of multi‐token (MT) fusion methods and number of tokens on Overall performance (\%). OR: Orthogonal Multi‐token Regularization.}
\label{tab:mt-fuse}
\begin{tabular}{@{}lrrrrrr@{}}
\toprule
\small
\multirow{2}{*}{\textbf{MT Fuse Method}} & \multicolumn{6}{c}{\textbf{Num of tokens}} \\ 
\cmidrule(l){2-7}
                                          & \textbf{1}  & \textbf{4}  & \textbf{8}  & \textbf{16} & \textbf{32} & \textbf{64} \\
\midrule
w/o MT ([EOS])                           & 67.9        & —           & —           & —           & —           & —           \\
Avg                                     & —           & 68.6        & 68.4        & 68.4        & 68.2        & 68.5        \\
Avg + OR                                & —           & 68.3        & 68.3        & 68.7        & 68.7        & \bfseries 68.8 \\
\bottomrule
\end{tabular}
\end{table}

To illustrate MEF’s advantage in retaining pre‐trained knowledge, Figure \ref{fig:multi-token} presents two VQA cases. In Place365, our model (Exp 6) correctly retrieves the “science\_museum” scene. In ScienceQA retrieval (“What is the capital of Mississippi?”), it retrieves “Jackson,” whereas the [EOS]-based baseline fails. By distributing information across multiple learnable tokens rather than condensing into a single [EOS] vector, MEF preserves the MLLM’s internal representations and better leverages its world knowledge. We also include in the appendix a visual comparison between average fusion and our orthogonal multi-token fusion, demonstrating the necessity of the orthogonality constraint.

\section{Conclusion}
We present ReMatch, a simple yet effective framework that unifies generative and discriminative representations in MLLMs for retrieval tasks. 
By introducing a chat-style generative matching strategy, Rematch embraces the autoregressive paradigm to deliver instance‐level, multi-view relevance judgments while mitigating hard‐negative bias.
Moreover, we design learnable multi-token augmentation to produce fine-grained, mutually orthogonal embeddings, thereby better exploiting the MLLM’s world knowledge.
Extensive experiments show that our approach achieves SOTA on MMEB benchmark with remarkable zero-shot transfer across five retrieval tasks. ReMatch paves the way for a deeper unification of generative and discriminative paradigms in future multi-modal models.



{
    \small
    \bibliographystyle{ieeenat_fullname}
    \bibliography{main}
}
\clearpage
\setcounter{page}{1}
\renewcommand{\thesection}{S\arabic{section}}
\maketitlesupplementary
\definecolor{myblux}{rgb}{0.113,0.65,0.78}

\section{Embedding Distribution Visualization}
\begin{figure*}[t]
\begin{center}
\includegraphics[width=1\linewidth]{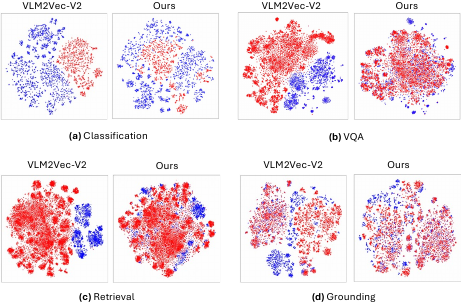}
\end{center}
\caption{t-SNE (perplexity = 30) of \textcolor{red}{query} and \textcolor{blue}{document} embeddings on four tasks, comparing VLM2Vec-V2 2B (left) with our ReMatch 2B (Qwen2-VL, right). Desired geometry is tight query–document co-localization in compact, overlapping clusters. Conversely, pronounced red–blue segregation or large single-color regions indicate weak query–document alignment and degraded embedding quality. Best viewed with zoom.}
\label{fig:tsne}
\end{figure*}

We first assess whether queries and documents are semantically aligned, i.e., embedded in close proximity without domain shift.
Figure \ref{fig:tsne} presents t-SNE visualizations of query–document embeddings on the MMEB benchmark, comparing the baseline VLM2Vec-V2 with our ReMatch. For each of the four tasks, we randomly sample three different subtasks. On all settings, ReMatch yields substantially tighter clustering between matched queries and documents than VLM2Vec-V2, reflecting its improved embedding alignment. 
Notably, on the VQA and retrieval tasks, VLM2Vec-V2 produces two clearly separated clusters, one with queries and the other with documents; in contrast ReMatch achieves a great increase in distribution overlap.

\section{Full Prompt for MVQDM}

This is our full prompt template for MVQDM with eight views. Due to space limits in the main text, we only present a simplified version in Sec.~\ref{sec:qdm}.

\begin{minipage}{1.0\columnwidth}\vspace{2mm}\hspace{-8mm}    \centering
\begin{adjustbox}{width=1.0\columnwidth}
\begin{tcolorbox}
    \raggedright
    \small
{
\textbf{Prompt Template for MVQDM with Eight Views:}\\

\texttt{SYSTEM: 
You are a helpful assistant.} 

\texttt{USER: Compare the content inside }
\PredStyOrange{<QUERY>}\texttt{ and }\texttt{\textcolor{magenta}{<DOC>}}\texttt{. Determine whether the information in }\texttt{\textcolor{magenta}{<DOC>}}\texttt{ is relevant to the query in }\PredStyOrange{<QUERY>}\texttt{. Answer only with "Yes" or "No", and wrap your answer in }\texttt{\textcolor{myblux}{<ANSWER></ANSWER>}}\texttt{ tags.}

\texttt{\PredStyOrange{<QUERY>}}
\texttt{\PredSty{\{query\}}}
\texttt{\PredStyOrange{</QUERY>} } \\
\texttt{\PredStyOrange{<QUERY>}}
\texttt{<feat\_start>}
\texttt{\PredSty{\{query\_feat\}}}
\texttt{<feat\_end>}
\texttt{\PredStyOrange{</QUERY>} }\\

\texttt{\textcolor{magenta}{<DOC>}}
\texttt{\PredSty{\{doc1\}}}
\texttt{\textcolor{magenta}{</DOC>} }\\
\texttt{\textcolor{magenta}{<DOC>}}
\texttt{<feat\_start>}
\texttt{\PredSty{\{doc1\_feat\}}}
\texttt{<feat\_end>}
\texttt{\textcolor{magenta}{</DOC>} } \\

\texttt{\textcolor{magenta}{<DOC>}}
\texttt{\PredSty{\{doc2\}}}
\texttt{\textcolor{magenta}{</DOC>}}  \\
\texttt{\textcolor{magenta}{<DOC>}}
\texttt{<feat\_start>}
\texttt{\PredSty{\{doc2\_feat\}}}
\texttt{<feat\_end>}
\texttt{\textcolor{magenta}{</DOC>}} 

\texttt{ASISTANT: }
\texttt{\textcolor{myblux}{<ANSWER>}\PredSty{\{ans1\}}\textcolor{myblux}{</ANSWER>}}
\texttt{\textcolor{myblux}{<ANSWER>}\PredSty{\{ans1\}}\textcolor{myblux}{</ANSWER>}}
\texttt{\textcolor{myblux}{<ANSWER>}\PredSty{\{ans2\}}\textcolor{myblux}{</ANSWER>}}
\texttt{\textcolor{myblux}{<ANSWER>}\PredSty{\{ans2\}}\textcolor{myblux}{</ANSWER>}}
\texttt{\textcolor{myblux}{<ANSWER>}\PredSty{\{ans1\}}\textcolor{myblux}{</ANSWER>}}
\texttt{\textcolor{myblux}{<ANSWER>}\PredSty{\{ans1\}}\textcolor{myblux}{</ANSWER>}}
\texttt{\textcolor{myblux}{<ANSWER>}\PredSty{\{ans2\}}\textcolor{myblux}{</ANSWER>}}
\texttt{\textcolor{myblux}{<ANSWER>}\PredSty{\{ans2\}}\textcolor{myblux}{</ANSWER>}}
}

\vspace{-1mm}
\end{tcolorbox}
\end{adjustbox}
\vspace{1mm}
\end{minipage}

Moreover, we introduce MVQDM\textsuperscript{++}, which embeds the inputs within a chat-style template to fully align with the matching paradigm. By contrast, previous methods, such as VLM2Vec-V2 or B3, derive embeddings from inputs of the form:
\texttt{\{image\} \{text\} \{task instruction\}}.

\begin{minipage}{1.0\columnwidth}\vspace{2mm}\hspace{-7mm}    \centering
\begin{adjustbox}{width=1.0\columnwidth}
\begin{tcolorbox}
    \raggedright
    \small
    
\textbf{Embedding Prompt Template for MVQDM\textsuperscript{++}:}\\
\texttt{SYSTEM: 
You are a helpful assistant.} 
\texttt{USER:\{image\} \{text\} \{task instruction\}} \\

\texttt{ASSISTANT:}

\vspace{-1mm}
\end{tcolorbox}
\end{adjustbox}
\vspace{1mm}
\end{minipage}

\section{Effectiveness of orthogonal regularization in multi-embedding fusion.}

\begin{figure*}[t]
\begin{center}
\includegraphics[width=0.92\linewidth]{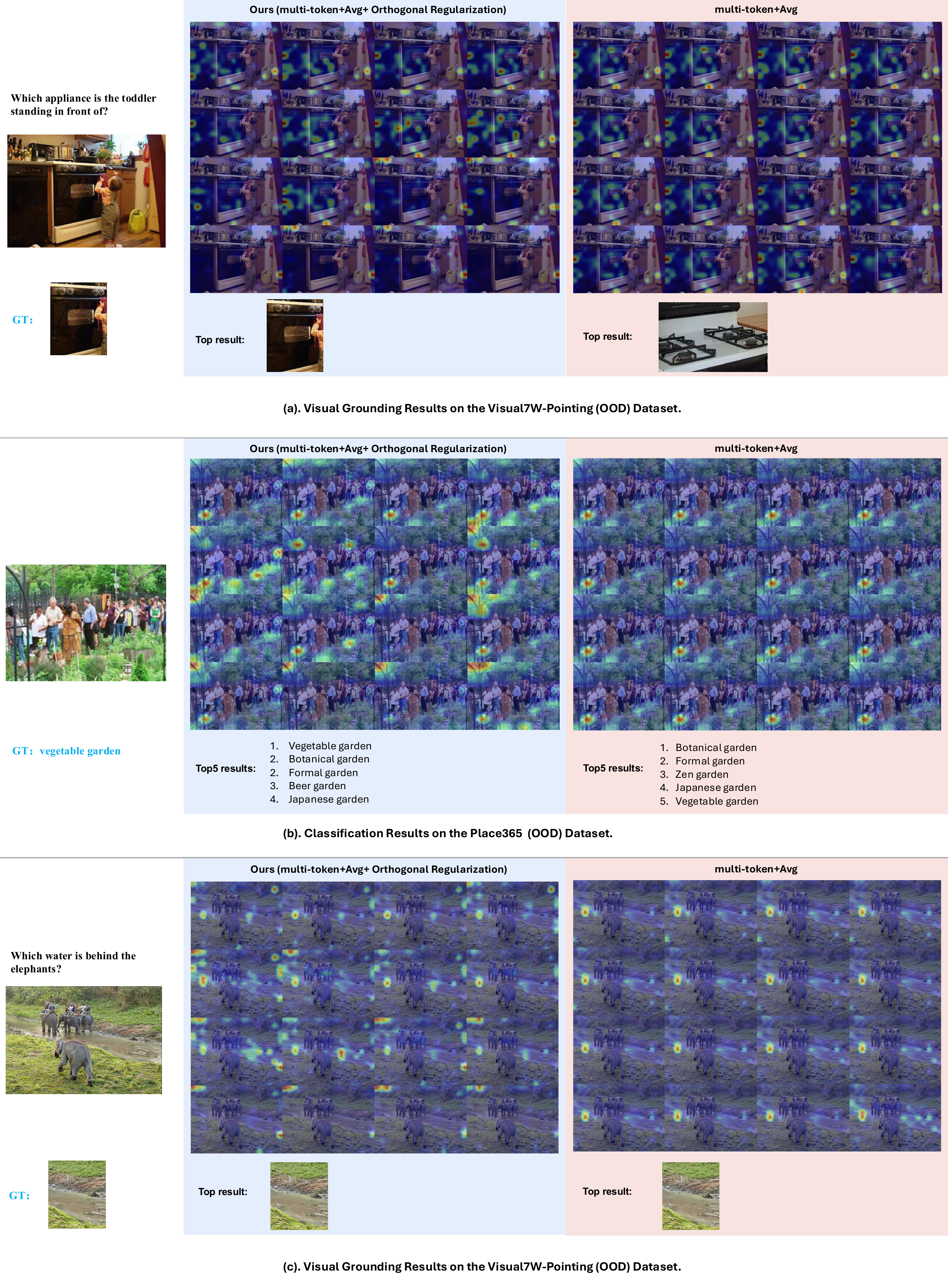}
\end{center}
\vspace{-10pt}
\caption{Attention maps between the query image and each learnable token (16 tokens; 4×4 grid) and retrieval results of our method, with and without orthogonal regularization in MEF, across three cases.}
\label{fig:att_mt}
\end{figure*}

To produce the attention visualizations in Figures \ref{fig:motivation}, \ref{fig:match_figure} and \ref{fig:att_mt}, we extract the MLLM’s last layer ($28^{th}$ of Qwen-VL-2B) attention weights between each embedding token (VLM2Vec-V2 uses last token; our method includes all learnable tokens) and the image patches. For each token, we average the attention weights across heads, apply a min-max normalization to [0,1], and reshape it into the original patch‐grid dimensions. We then overlay this attention map onto the RGB input. Side by side, these overlays reveal the different regions each token focuses on, showing the richer, more diverse semantics of our multi-token representations.

Figure~\ref{fig:att_mt} validates our key hypothesis: simply appending multiple learnable tokens (MEF + Avg) often yields highly redundant information across tokens. Each token attends to the same image regions, thus increasing the token count alone does not improve performance. By contrast, adding our soft orthogonal regularization (MEF + Avg + OR) forces each token to specialize on different parts of the scene, enriching the overall embedding. 
Specifically, in case (a), only the OR-regularized model correctly grounds the appliance that the toddler is standing in front of the oven; in case (b), the regularized version attends separately to the upper and lower tree clusters and correctly classifies the scene as a “vegetable garden,” whereas the unregularized maps collapse onto the lower foliage and misclassify it; and in case (c), the attention maps without OR look virtually identical and still miss large portions of the riverbank, whereas our OR-enhanced tokens collectively cover the full watercourse and recover the correct grounding.

\section{Beyond Embeddings: Can Embedding MLLMs Still Speak Fluently?}
\begin{figure*}[t]
\begin{center}
\includegraphics[width=0.92\linewidth]{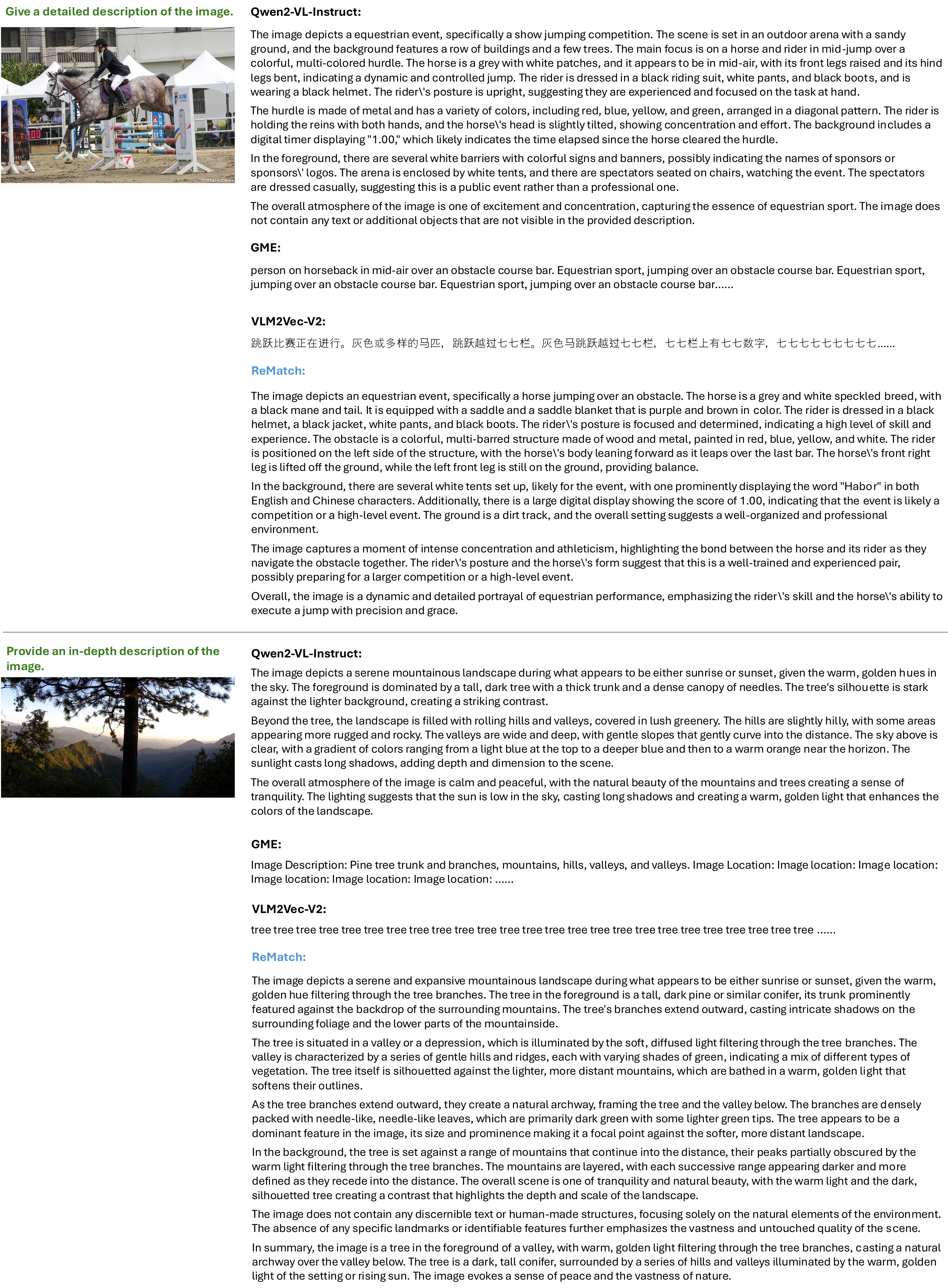}
\end{center}
\vspace{-10pt}
\caption{Comparison of image captioning performance between our method, original backbone and prior embedding approaches (all based on Qwen2-VL-2B). Part 1.}
\label{fig:gene_task_1}
\end{figure*}

\begin{figure*}[t]
\begin{center}
\includegraphics[width=1\linewidth]{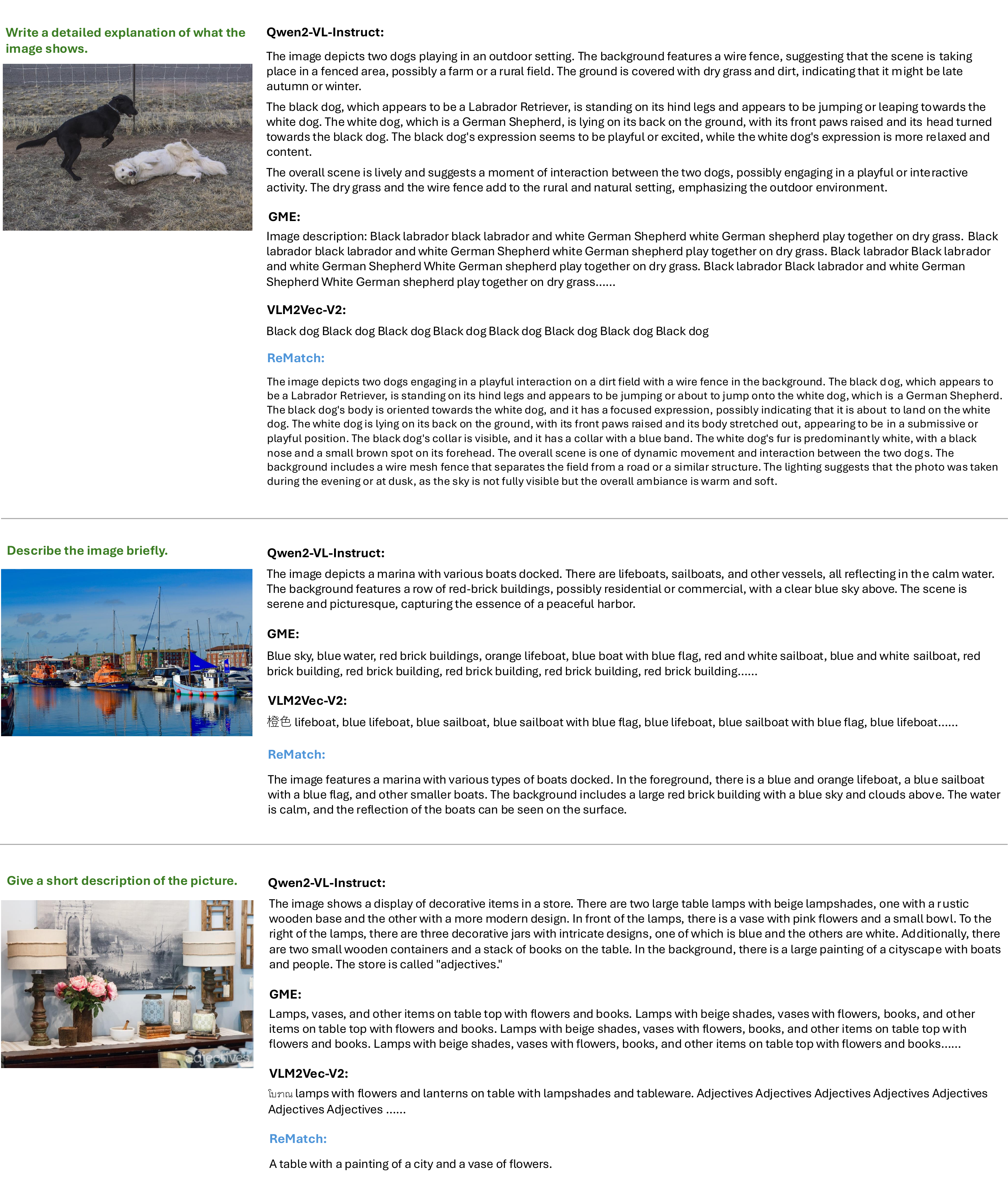}
\end{center}
\vspace{-10pt}
\caption{Comparison of image captioning performance between our method, original backbone and prior embedding approaches (all based on Qwen2-VL-2B). Part 2.}
\label{fig:gene_task_2}
\end{figure*}

As discussed above, multimodal embedding models disregard the backbone’s autoregressive nature. They improve discriminative ability via contrastive learning but may attenuate the backbone’s world knowledge and generative capacity. This prompts the question: after contrastive fine-tuning, can these embedding models still preserve image-captioning performance? Here, unlike elsewhere in the paper, the MLLM generates text autoregressively, conditioned on the image provided in its prompt.
Figure~\ref{fig:gene_task_1} and \ref{fig:gene_task_2} present results on three long-caption and two short-caption cases, comparing the original backbone (Qwen2-VL-2B) with two representative embedding methods— GME (compresses features into [EOS]) and VLM2Vec-V2 (uses the last token). Across all five examples, both baselines fail to produce fluent descriptions, frequently exhibiting repetition and VLM2Vec-V2 even shows mixed-language outputs (e.g., Chinese and Thai). In contrast, ReMatch nearly seamlessly retains the comparable generation quality of Qwen2-VL-2B. 
This finding indicates that single-token embedding compression at [EOS] or at the last token degrades the backbone’s autoregressive generation ability. By contrast, a learnable multi-token enhancement combined with a chat style matching strategy not only substantially improves embedding performance but also preserves the model’s generative capacity.

\section{More Visualization for four tasks on MMEB}

\begin{figure*}[t]
    \centering %

    \includegraphics[width=0.9\linewidth]{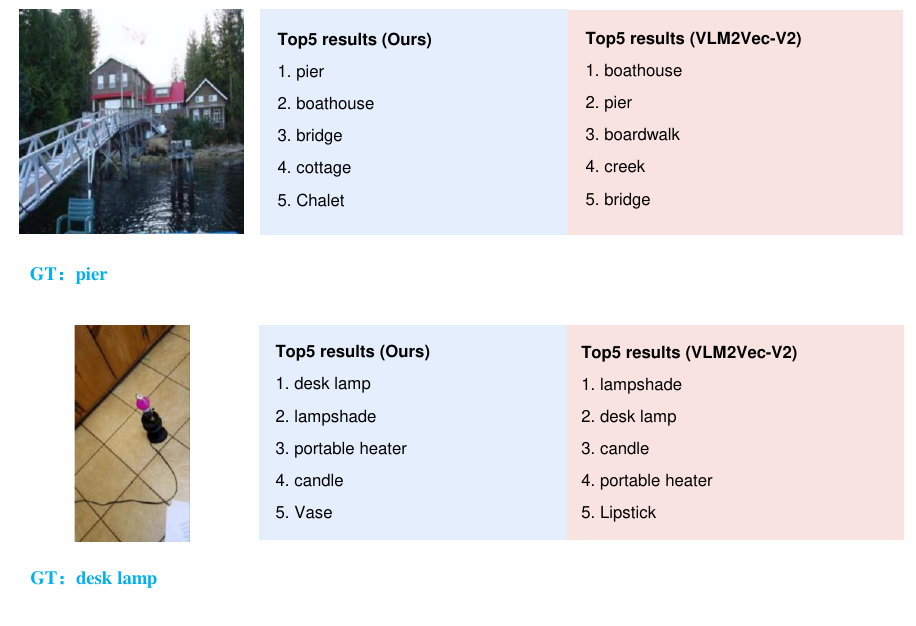}
    \vspace{-10pt}
    \caption{Comparison of classification performance between ReMatch and the baseline.}
    \label{fig:MMEB1}
    
    \vspace{10pt} %
    \includegraphics[width=0.9\linewidth]{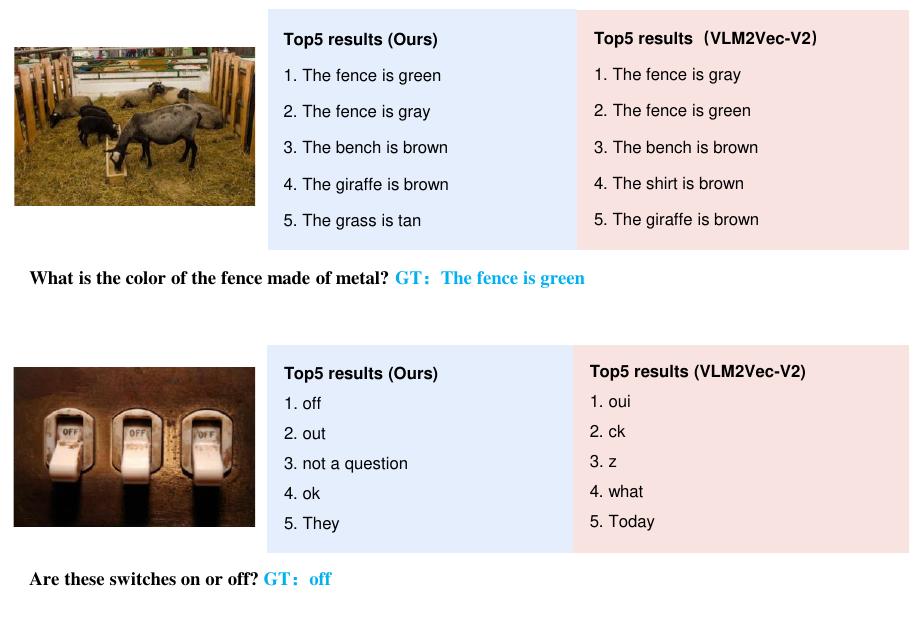}
    \vspace{-10pt}
    \caption{Comparison of visual question answering performance between ReMatch and the baseline.}
    \label{fig:MMEB2}
    
\end{figure*}

\begin{figure*}[t]
\begin{center}
\includegraphics[width=0.9\linewidth]{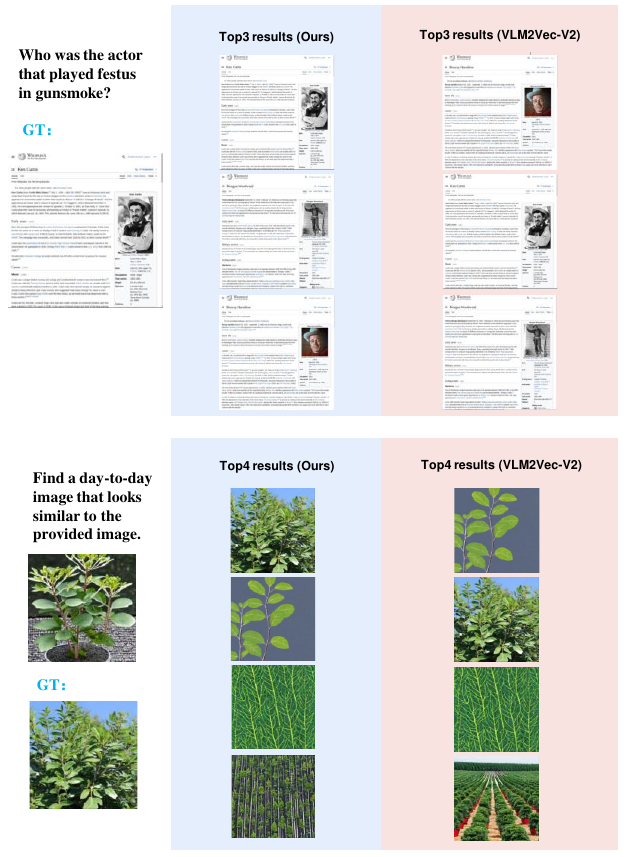}
\end{center}
\vspace{-10pt}
\caption{Comparison of retrieval performance between ReMatch and the baseline. Confidence-ordered from top to bottom (highest to lowest).}
\label{fig:MMEB3}
\end{figure*}

\begin{figure*}[t]
\begin{center}
\includegraphics[width=0.9\linewidth]{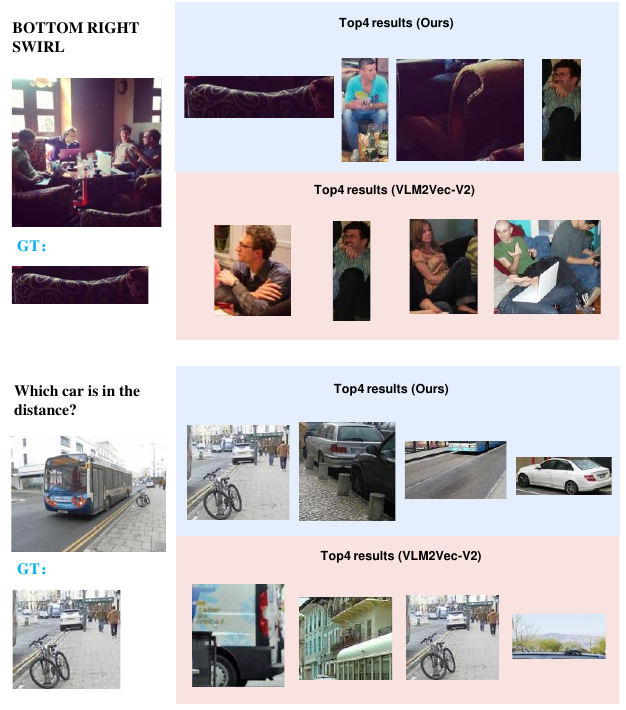}
\end{center}
\vspace{-10pt}
\caption{Comparison of visual grounding performance between ReMatch and the baseline. Confidence-ordered from left to right (highest to lowest).}
\label{fig:MMEB4}
\end{figure*}

In this section, we present additional visual comparisons between our method and VLM2Vec-V2 on the MMEB benchmark (Figures~\ref{fig:MMEB1}–\ref{fig:MMEB4}). For each of four tasks (classification, VQA, retrieval, and visual grounding), we report two cases. In the resulting eight examples, our method shows consistently superior performance.

\noindent \textbf{Note.} In the main text, we use VLM2Vec as our baseline (the starting point for ReMatch), while in the appendix we compare against a stronger variant, VLM2Vec-V2, to further emphasize our performance gains.

\end{document}